\documentclass[10pt,twocolumn,letterpaper]{article}

\usepackage[numbers]{natbib}
\usepackage{cvpr}
\usepackage{times}
\usepackage{epsfig}
\usepackage{graphicx}
\usepackage{amsmath}
\usepackage{amssymb}
\usepackage{multirow}
\usepackage{bm}
\usepackage{booktabs}
\usepackage{subfig}
\usepackage[colorlinks,linkcolor=blue]{hyperref}

\cvprfinalcopy 

\def\cvprPaperID{****} 

\ifcvprfinal\fi
\begin{document}

\title{OPA: Object Placement Assessment Dataset}

\author{$\textnormal{Liu Liu}^{*}$,
$\textnormal{Zhenchen Liu}^{*}$,
$\textnormal{Bo Zhang}^{*}$, $\textnormal{Jiangtong Li}^{*}$, $\textnormal{Li Niu}^{*}$, $\textnormal{Qingyang Liu}^{\dag}$, $\textnormal{Liqing Zhang}^{*}$\\
$^*$ Shanghai Jiao Tong University\,\,
$\dag$ Beijing Institute of Technology \\
}

\maketitle
\thispagestyle{empty}

\begin{abstract}
Image composition aims to generate realistic composite image by inserting an object from one image into another background image, where the placement (\emph{e.g.}, location, size, occlusion) of inserted object may be unreasonable, which would significantly degrade the quality of the composite image.
Although some works attempted to learn object placement to create realistic composite images, 
they did not focus on assessing the plausibility of object placement. 
In this paper, we focus on object placement assessment task, which verifies whether a composite image is plausible in terms of the object placement. 
To accomplish this task, we construct the first Object Placement Assessment (OPA) dataset consisting of composite images and their rationality labels. We also propose a simple yet effective baseline for this task. 
 Dataset is available at \href{https://github.com/bcmi/Object-Placement-Assessment-Dataset-OPA}{https://github.com/bcmi/Object-Placement-Assessment-Dataset-OPA}.
\end{abstract}


\section{Introduction}
As a common image editing operation, image composition aims to generate a realistic-looking image by pasting the foreground object of one image on another image.
The composites can result in fantastic images that previously only exist in the imagination of artists, which can greatly benefit a variety of applications like augmented reality and artistic creation \cite{niu2021making,WhereandWhoTan2018,MISCWeng2020}.
However, it is challenging to insert a foreground object into a background image that satisfies the following requirements:
1) the foreground object has compatible color and illumination with the background image;
2) the inserted object may have an impact on the background image, like the reflection and shadow;
3) the foreground object should be placed at a reasonable location on the background considering location, size, occlusion, semantics, and \emph{etc}.
To satisfy the above requirements, image harmonization \cite{TsaiDIHarmonization2017, CongDoveNet2020}, shadow generation \cite{ExposingKee2014, ARShadowGANLiu2020}, and object placement \cite{WhereandWhoTan2018, STGANChen2018} have been proposed to improve the quality of composite images from the above aspects, respectively.

\begin{figure*}
\centering
\includegraphics[width=1.\linewidth]{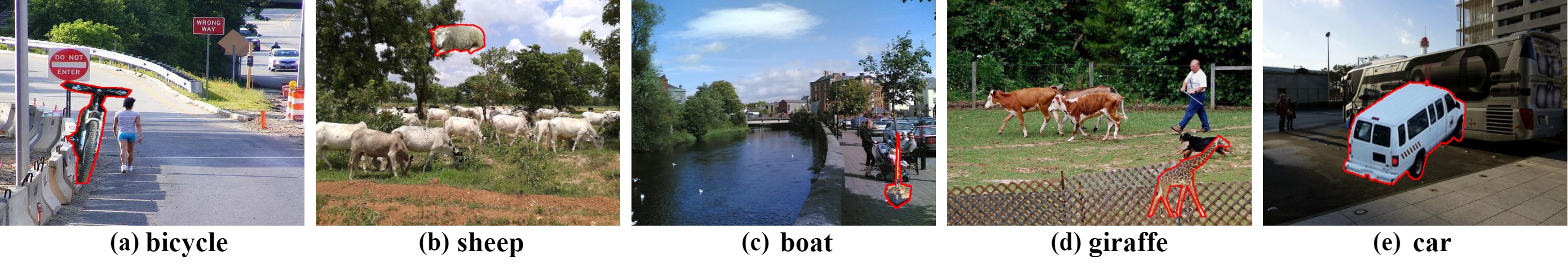}
\caption{Some negative samples in our OPA dataset and the inserted foreground objects are marked with red outlines. From left to right: (a) objects with inappropriate size; (b) objects hanging in the air; (c) objects appearing in the semantically unreasonable place; (d) unreasonable occlusion; (e) inconsistent perspectives. }
\label{fig:dataset}
\end{figure*}

In this paper, we focus on the third issue, object placement, aiming to paste foreground object on the background with suitable location, size, occlusion, \emph{etc}. As shown in Figure \ref{fig:dataset}, the cases of unreasonable object placement \cite{CompositionalGANAzadi2020} are including but not limited to: 1) the foreground object is too large or too small;
2) the foreground object does not have supporting force (\emph{e.g.}, hanging in the air); 3) the foreground object appears in a semantically unreasonable place (\emph{e.g.}, boat on the land);
4) unreasonable occlusion;
5) inconsistent perspectives between foreground and background.
The above unreasonable cases would significantly degrade the reality of composite images.
Considering a wide range of foreground objects and complicated scenarios, object placement is still a challenging task.

Some previous works attempted to learn reasonable object placement to generate realistic composite images. One group of methods \cite{STDODISGeorgakis2017, LSCPRemez2018, InstanceSwitchingWang2019, InstaBoostFang2019} relied on explicit rules to find a reasonable location for the foreground object. For example, the new background of inserted foreground should be close to its original background \cite{InstaBoostFang2019} or the foreground should be placed on a flat plane \cite{STDODISGeorgakis2017}. 
However, these explicit rules are only applicable to limited scenarios. The other group of methods trained network to automatically learn the reasonable object placement, which can be further divided into supervised and unsupervised methods.  
Supervised methods \cite{WhereandWhoTan2018, MVCIKAODDDvornik2018, WhatWhereZhang2020, LearningObjPlaZhang2020, ContextawareLee2018} leveraged the size/location of foreground object in the original image as ground-truth. They predicted the bounding box or transformation of the foreground object based on the foreground and background features \cite{WhereandWhoTan2018,LearningObjPlaZhang2020}.
Unsupervised methods like 
\cite{SyntheticTripathi2019} did not use ground-truth size/location. They learned reasonable transformation of foreground object, by pushing the generated composite images close to real images. 

All the above works focus on generating reasonable composite images instead of object placement assessment. In other words, they cannot automatically assess the rationality of a composite image in terms of object placement.  
To evaluate the quality of generated composite images, the above works on learning object placement usually adopt the following three approaches. 1) \cite{WhereandWhoTan2018} scored the correlation between the distributions of predicted boxes and ground-truth boxes. \cite{LearningObjPlaZhang2020} calculated the Frechet Inception Distance (FID) \cite{FIDHeusel2017} between composite and real images to measure the placement plausibility.
However, they cannot evaluate each individual composite image.  2) \cite{SyntheticTripathi2019, InstaBoostFang2019} utilized the improvement of downstream tasks (\emph{e.g.}, object detection) to evaluate the quality of composite images, where the training sets of the downstream tasks are augmented with generated composite images.
However, the evaluation cost is quite huge and the improvement in downstream tasks may not reliably reflect the quality of composite images, because \cite{SimpleCopyPasteGhiasi2020} revealed that randomly generated composite images could also boost the performance of downstream tasks. 
3) Another common evaluation strategy is user study, where people are asked to score the rationality of placement \cite{ContextawareLee2018, WhereandWhoTan2018}. User study complies with human perception and each composite image can be evaluated individually. However, due to the subjectivity of user study, the gauge in different papers may be dramatically different. There is no unified benchmark dataset and the results in different papers cannot be directly compared. 


In summary, as far as we are concerned, no previous works focus on object placement assessment and no suitable dataset is available for this task. In this work, we focus on the task of object placement assessment, that is, automatically assessing the rationality of a composite image in terms of object placement. 
We build an Object Placement Assessment (OPA) dataset for this task, based on COCO \cite{MSCOCOLin2014} dataset.
First, we select unoccluded objects from multiple categories as our candidate foreground objects.
Then, we design a strategy to select compatible background images for each foreground object.
The foreground objects are pasted on their compatible background images with random sizes and locations to form composite images, which are sent to human annotators for rationality labeling.
Each image is labeled by four human annotators, where only the images with consistent labels are preserved in the dataset to ensure the annotation quality.
Finally, we split the collected dataset into training set and test set, in which the background images and foreground objects have no overlap between training set and test set.
More details about constructing the dataset will be elaborated in Section \ref{sec:data-construction}.

With the constructed dataset, we regard the object placement assessment task as a binary classification problem and any typical classification network can be applied to this task. 
We propose a simple yet effective baseline for object placement assessment task by feeding the concatenation of composite image and foreground mask into a classification network (\emph{e.g.}, ResNet~\cite{he2015resnet}). 
With the functionality of object placement assessment, our model can help obtain realistic composite images. Particularly, given automatically (\emph{e.g.}, \cite{SyntheticTripathi2019, LearningObjPlaZhang2020}) or manually (\emph{e.g.}, by users) created composite images, we can apply object placement assessment model to select the composite images with high rationality scores. Our model may also assist designers during artistic creation, by providing feedback and making recommendation in terms of object placement. 
Our main contributions can be summarized as follows,
\begin{itemize}
\item We propose a new task named object placement assessment (OPA), which aims to automatically assess the rationality of a composite image in terms of object placement.
\item We construct and release the first object placement assessment (OPA) dataset, which is comprised of composite images and their binary rationality labels. 
\item We propose a simple yet effective baseline named SimOPA for OPA task, which can achieve reasonable results on our OPA dataset. 
\end{itemize}

\section{Dataset Construction} \label{sec:data-construction}
In this section, we describe the construction process of our Object Placement Asssessment (OPA) dataset, in which we first generate composite images and then ask human annotators to label these composite images \emph{w.r.t.} the rationality of object placement. 

\subsection{Composite Image Generation} \label{subsec: composite image generation}
We select suitable foreground objects and background images from Microsoft COCO dataset \cite{MSCOCOLin2014}, which are used to generate composite images.

\noindent\textbf{Foreground object selection:}
There are 80 object categories in COCO \cite{MSCOCOLin2014}  with annotated instance segmentation masks. We only keep unoccluded foreground objects, because it is difficult to find reasonable placement for occluded objects. 
We delete some categories according to the following rules: 1) the categories which usually appear at very specific locations, such as transportation-related categories (\emph{e.g.}, traffic light, stop sign) and human-centric categories (\emph{e.g.}, tie, snowboard);
2) the categories of large objects appearing in crowded space, such as large furniture (\emph{e.g.}, refrigerator, bed);
3) the categories with too few remaining objects after removing occluded and tiny foreground objects (\emph{e.g.}, toaster, hair drier); 
4) the categories with high disagreement among multiple
annotators in the pilot study (\emph{e.g.}, handbag). In summary, the above categories are either hard to find reasonable placement or hard to verify the rationality of object placement. 
After filtering, 47 categories remain.

The complete list of 47 categories is: \emph{airplane, apple, banana, bear, bench, bicycle, bird, boat, book, bottle, bowl, broccoli, bus, cake, car, cat, cellphone, chair, cow, cup, dog, donut, elephant, fire hydrant, fork, giraffe, horse, keyboard, knife, laptop, motorcycle, mouse, orange, person, pizza, potted plant, remote, sandwich, scissors, sheep, spoon, suitcase, toothbrush, truck, vase, wineglass, zebra.}

With the annotated instance segmentation masks from COCO \cite{MSCOCOLin2014} dataset, we select 100 unoccluded foreground objects for each category.

\noindent\textbf{Background image selection:}
For each foreground category, there should be a set of compatible background images. For example, airplanes do not appear indoors and forks usually appear on the table. In this work, we eliminate the burden of selecting compatible background images for object placement assessment task.

We fine-tune PlaceCNN \cite{placesCNNZhou2017} pretrained on places365 \cite{placesCNNZhou2017} to select a set of compatible background images for each category.
Specifically, for each category, we take the images containing the objects of this category as positive samples, and randomly sample an equal number of other images as negative samples. Then, we fine-tune PlaceCNN \cite{placesCNNZhou2017} based on positive and negative samples to learn a binary classifier. For each category, we apply the trained binary classifier to retrieve top $100$ images which do not contain the objects of this category as a set of compatible background images. We set this rule to prevent the shortcut
that the learnt model directly utilizes the clues of the background object from
the same category.

\noindent\textbf{Composite image generation:}
We generate a composite image by pasting one foreground object on another background image.
To avoid too much repetition, we limit the size and location of the foreground object according to some prior knowledge.

For each foreground category, we first calculate a reasonable range of its size ratio, which is defined as the ratio of foreground object size over its corresponding image size.  
Given a foreground object and a compatible background image, we randomly sample $5$ size ratios and $9$ locations, leading to $45$ composite images. For size ratio, we divide the range of size ratio of foreground category into five bins based on  20\%, 40\%, 60\%,  80\% quantiles, and randomly sample one size ratio from each bin. For location, we evenly divide the background image into $9$ partitions and randomly sample one location from each partition. We resize the foreground object according to certain size ratio and place it at certain location, producing a composite image. 
Besides, we remove the composite images with incomplete foreground objects (\emph{e.g.}, half of the foreground object is out of the scope of the background image), because the incomplete information raises the difficulty of human annotation in practice.

\begin{figure*}
\centering
\includegraphics[width=1.\linewidth]{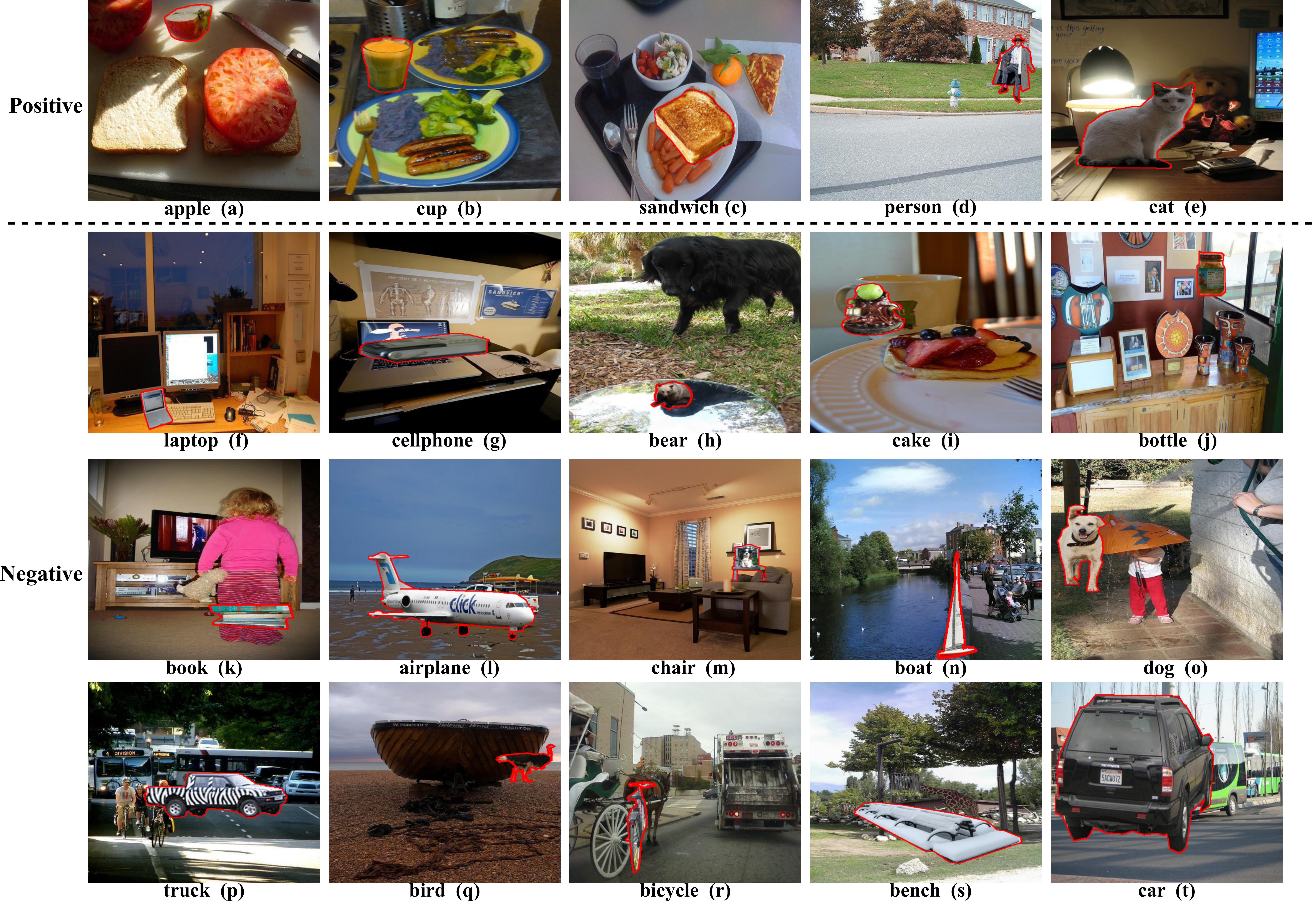}
\caption{Some positive and negative samples in our OPA dataset and the inserted foreground objects are marked with red outlines. Top row: positive samples; Bottom rows: negative samples, including objects with inappropriate size (\emph{e.g.}, f, g, h), without supporting force (\emph{e.g.}, i, j, k), appearing in the semantically unreasonable place (\emph{e.g.}, l, m, n), with unreasonable occlusion (\emph{e.g.}, o, p, q), and with inconsistent perspectives (\emph{e.g.}, r, s, t).}
\label{fig:datset_egs}
\end{figure*}

\begin{figure*}
\centering
\includegraphics[width=0.9\linewidth]{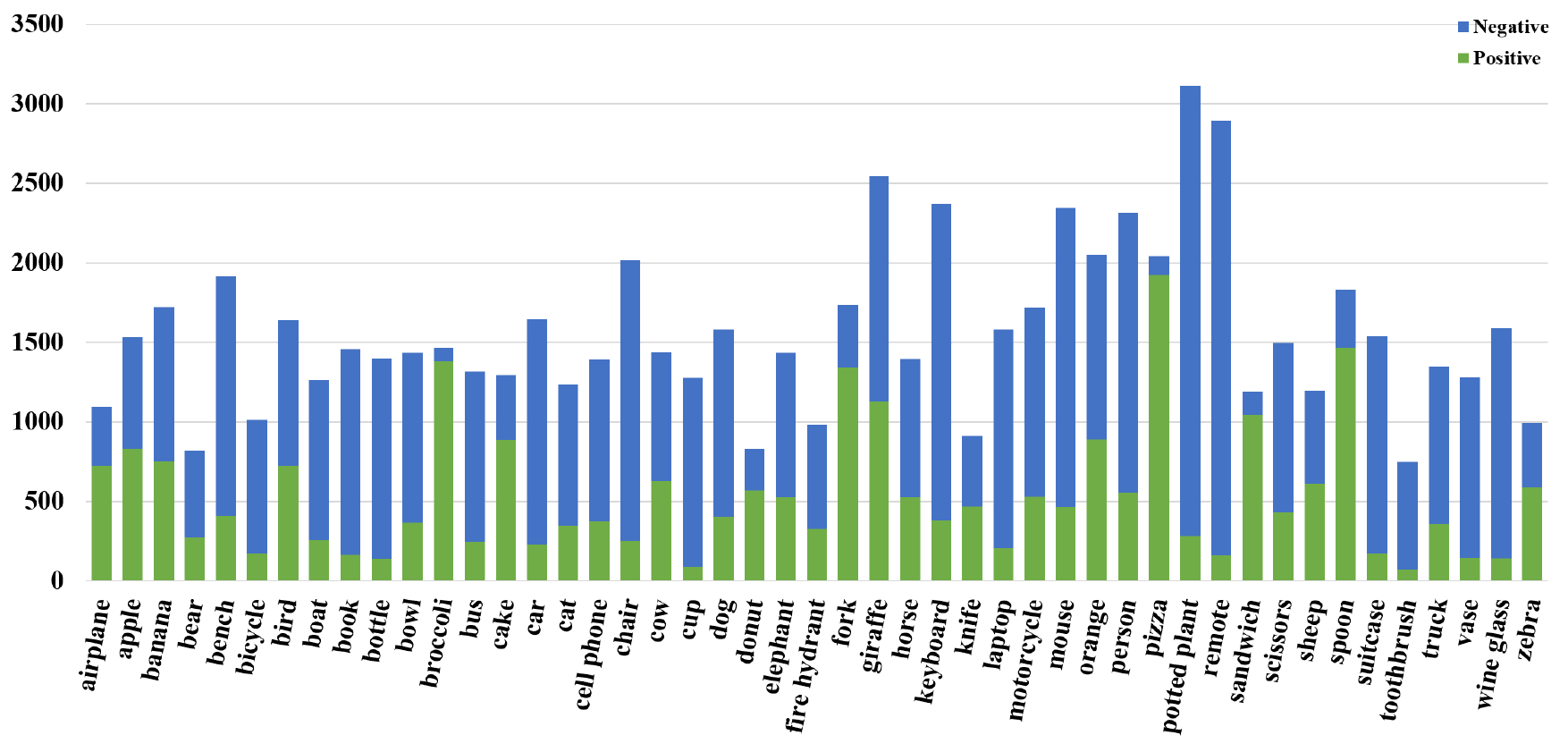}
\caption{The number of images per foreground category in our OPA dataset.}
\label{fig:datset_nums}
\end{figure*}

\begin{figure*}[t]
\centering
\includegraphics[width=0.9\linewidth]{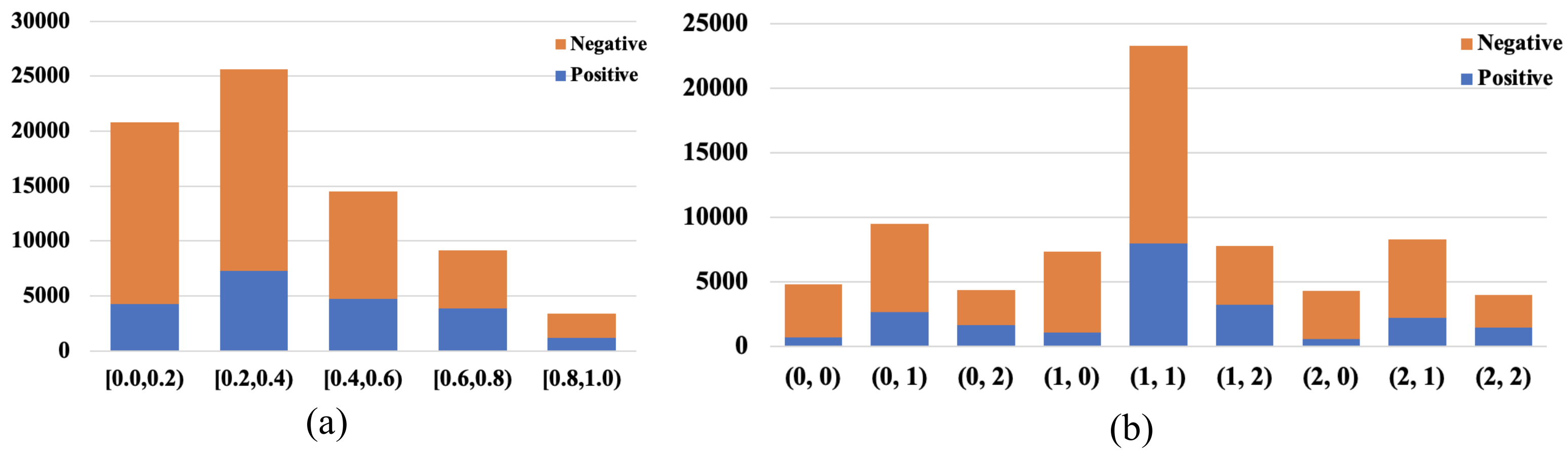}
\caption{The number of images (positive and negative) per size ratio range (a) and per location partition (b) in our OPA dataset. In (b), $(i,j)$ indicates the partition in the $i$-row and $j$-th column}
\label{fig:datset_histogram}
\end{figure*}

\subsection{Composite Image Labelling}
Since the rationality of object placement is constrained by many complicated factors (\emph{e.g.}, location, size, occlusion, semantics), the number of negative images is significantly larger than the positive samples among the randomly generated composite images.
To achieve relatively balanced positive-negative ratio and save the human labor, we first fine-tune a ResNet-50 \cite{he2015resnet}  classifier pretrained on ImageNet \cite{ImageNet2009} to remove the obviously unreasonable composite images.
During fine-tuning, the real images are regarded as positive samples. We additionally generate composite images via random copy-and-paste as negative samples, which have no overlap with the composite images in Section~\ref{subsec: composite image generation}.  Although the generated composite images contain both positive samples and negative samples, negative samples are dominant and thus the learned binary classifier is useful. 
To indicate the foreground object, we also feed foreground mask into ResNet-50 \cite{he2015resnet} classifier.
We apply the fine-tuned classifier to the composite images in Section~\ref{subsec: composite image generation} and select the top 235,000 composite images with the highest scores for further labeling. The selected composite images are supposed to have relatively higher ratio of positive samples. 

To acquire the binary rationality label ($1$ for reasonable object placement and $0$ for unreasonable object placement), we ask four human annotators to label the rationality for each composite image.
We purely focus on the object placement issues and ignore the other issues (\emph{e.g.}, inconsistent illumination between foreground and background, unnatural boundary between foreground and background).
Due to the subjectivity of this annotation task, we make detailed annotation guidelines (\emph{e.g.}, the reasonable range of sizes for each foreground category) and train human annotators for two weeks to make the annotations consistent across different annotators. The detailed annotation guidelines are as follows,
\begin{enumerate}
\item All foreground objects are considered as real objects instead of models or toys.
\item The foreground object placement conforms to the basic laws of physics. Except for the flying objects (\emph{e.g.}, airplane), all the other objects should have reasonable supporting force. 
\item The foreground object should appear in a semantically reasonable place. We also make some specific rules for the ambiguous cases. For example, for the container categories (\emph{e.g.}, bowl, bottle), we stipulate that they cannot be surrounded by fried dish. 
\item If there is occlusion between the foreground object and background object, the rationality of occlusion should be considered.
\item The size of the foreground object should be judged based on its location and relative distance to other background objects.
\item We provide a reasonable range of size for each category and the estimated size of the foreground should be within the range of its category. For animal categories (\emph{e.g.}, dog, sheep), we treat the sizes of animals of all ages (from baby animal to adult animal) as reasonable sizes. 
\item The perspective of foreground object should look reasonable. 
\item The inharmonious illumination and color, and unreasonable reflection and shadow are out of the scope of consideration. 
\end{enumerate}

\emph{Although some of the above rules may be arguable, which depends on the definition of rationality, our focus is making the annotation criterion as explicit as possible and the annotations across different images as consistent as possible, so that the constructed dataset is qualified for scientific study.}
Besides, similar categories are labeled by the same group of human annotators to further mitigate the inconsistency. 
Finally, we only keep the images for which four human annotators reach the agreement. Inspired by \cite{MSCOCOLin2014}, we also check the annotation quality by comparing to dedicated workers (co-authors of the paper). Each dedicated worker is assigned with a different set of 1000 images and asked to check their rationality labels. We find that the agreement rate is as high as 98\%, which proves that the overall annotation quality is substantially high.
From the remaining images, we construct training set with 62,074 images and test set with 11,396 images, whose foreground objects and background images have no overlap. We impose this constraint to better evaluate the generalization ability of different methods, because the foreground object and background image are generally out of the scope of training set in real-world applications.

Note that we did not include real images with specified foreground as positive samples due to the following concern. We expect all images in our dataset to be truly composite images. If real images (\emph{resp.}, negative composite images) are used as positive (\emph{resp.}, negative) samples, the model may distinguish positive and negative samples based on other clues (\emph{e.g.}, jagged boundary, incompatible illumination), which is against our motivation.

\subsection{Dataset Statistics} 

After composite image generation and composite image labelling, there are 24,964 positive samples and 48,506 negative samples in our OPA dataset. Our OPA dataset has 4,137 unrepeated foreground objects and 1,389 unrepeated background images. 
We show some example positive and negative images in our dataset examples in Figure \ref{fig:datset_egs}. We also present the number of images (positive and negative) per foreground category in Figure \ref{fig:datset_nums}. It can be seen that the positive ratios of different categories are quite different, which is caused by different properties of different categories. For example, some categories have a wide range of reasonable scales and reasonable locations to be placed, while other categories have a relatively narrow range, leading to different proportions of positive images.

Additionally, we also report the number of images (positive and negative) per size ratio range and per location partition (see Section \ref{subsec: composite image generation}) in Figure \ref{fig:datset_histogram}. One observation is that the size ratios of most images are within the
range of [0.0, 0.6]. Another observation is that more images are located in the
center partition (1, 1), which can be explained as follows. Since we only keep the
composite images with complete foreground objects (see Section \ref{subsec: composite image generation}), the foreground centers in the retained composite images have higher
chances to fall into the center partition (1, 1).

We divide our OPA dataset into 62,074 training images and 11,396 test images, in which the foregrounds/backgrounds in training set and test set have no overlap. The training (\emph{resp.}, test) set contains 21,376 (\emph{resp.}, 3,588) positive samples and 40,698 (\emph{resp.}, 7,808) negative samples. Besides, the training (\emph{resp.}, test) set contains 2,701 (\emph{resp.}, 1,436)  unrepeated foreground objects and 1,236 (\emph{resp.}, 153) unrepeated background images.

\section{Experiments}
In this section, we conduct experiments on our constructed OPA dataset. 
For evaluation metrics, we adopt F1-score and balanced accuracy, since object placement assessment is an unbalanced binary classification task.

Since we are the first work focusing on object placement assessment, there is no previous work specifically designed for object placement assessment. We first apply the basic ResNet-18 \cite{he2015resnet} classifier, which takes in a composite image and predicts its rationality label. Based on basic ResNet-18, we also try concatenating the foreground mask with the input composite image to emphasize the composite foreground, which is dubbed as SimOPA. 
Additionally, we compare with two recent works~\cite{LearningObjPlaZhang2020,SyntheticTripathi2019} on learning object placement. Both PlaceNet \cite{LearningObjPlaZhang2020} and TERSE \cite{SyntheticTripathi2019} are GAN-based methods with a generator and a discriminator, in which the discriminator is responsible for judging the reality of generated composite image. The discriminator of PlaceNet takes the foreground feature, background feature, and the location/size parameters of foreground as input. For TERSE, we remove the target network because there is no downstream task in our work. The discriminator of TERSE directly takes a composite image as input. We also replace the discriminator in TERSE with ResNet-18 and mark the results with *.  
For both methods~\cite{LearningObjPlaZhang2020,SyntheticTripathi2019}, we use the foregrounds and backgrounds in our training set as the inputs for generator. When updating the discriminator, we use the composite images produced from the generator as negative samples and the positive images in our training set as positive ones. After training, we apply the discriminator for testing.

The experimental results are summarized in Table \ref{tab:compare}. It can be seen that the discriminators in \cite{LearningObjPlaZhang2020,SyntheticTripathi2019} perform much worse than the basic ResNet-18 classifier, which demonstrates that the discriminator is mainly used to enhance the generator and cannot be directly applied to object placement assessment. We also observe that SimOPA outperforms ResNet-18 by a large margin, which shows the importance of including foreground mask as input. Therefore, SimOPA can serve as a simple yet effective baseline for OPA task.

\begin{table}
   
    \centering
    \normalsize
    \setlength{\tabcolsep}{1.0mm}{
    \begin{tabular}{c|cc}
    \hline
    Method & F1-score & Balanced Accuracy \\ \hline
    ResNet-18 \cite{he2015resnet} & 0.680 & 0.772 \\ \hline
    PlaceNet \cite{LearningObjPlaZhang2020} & 0.488 & 0.524 \\
    TERSE \cite{SyntheticTripathi2019} & 0.476 & 0.500 \\ 
    TERSE* \cite{SyntheticTripathi2019} & 0.355 & 0.539 \\ \hline
    SimOPA & \textbf{0.780} & \textbf{0.842} \\ \hline
    \end{tabular}}
    \caption{Comparison of different methods on our OPA dataset. * means replacing the discriminator with ResNet-18. Best results are denoted in boldface.}
     \label{tab:compare}
\end{table}

\section{Conclusion}
In this work, we focus on the object placement assessment task, which verifies the rationality of object placement in a composite image. 
To support this task, we have contributed an Object Placement Assessment (OPA) dataset. This dataset will facilitate the research in automatic object placement, which can automatically forecast the diverse and plausible placement of foreground object on the background image. We also propose a simple yet effective baseline for future research.

{\small
\bibliographystyle{ieee_fullname}
\bibliography{main.bbl}
}

\end{document}